\documentclass[letterpaper]{article}

\pdfoutput=1

\usepackage{aaai}
\usepackage{times}
\usepackage{helvet}
\usepackage{courier}
\usepackage{url}
\usepackage{graphicx}
\usepackage{subfigure}
\usepackage{listings}
\usepackage{color}
\usepackage{xspace}

\begin{document}

\nocopyright

\title{Action-based Character AI in Video-games with CogBots Architecture: \\ A Preliminary Report}

\author{Davide Aversa\\
Sapienza University of Rome \\
Rome, Italy \\
\url{davide.aversa@gmx.com}
\And
Stavros Vassos \\
Sapienza University of Rome, \\
Rome, Italy \\
 \url{vassos@dis.uniroma1.it}
}

\maketitle

\begin{abstract}
\begin{quote}

In this paper we propose an architecture for specifying the interaction of
non-player characters (NPCs) in the game-world in a way that abstracts common tasks in four main
conceptual components,  namely \emph{perception, deliberation, control,
action}. We argue that this architecture, inspired by AI research on
autonomous agents and robots, can offer a number of benefits in the form
of abstraction, modularity, re-usability and higher degrees of personalization 
for the behavior of each NPC. We also show how this architecture can be used 
to tackle a simple scenario related to the navigation of NPCs under incomplete 
information about the obstacles that may obstruct the various way-points in
the game, in a simple and effective way.
\end{quote}
\end{abstract}

\section{Introduction}

An important part in the design of non-player characters (NPCs) in video-games has to do 
with the artificial intelligence (AI) of the characters in terms of their actions in the 
game. This amounts to handling a variety of problems that also involve low-level aspects 
of the game, including for example pathfinding, detection of conditions, execution 
monitoring for the actions of the characters, and many more. 

While many of these aspects have been studied extensively separately, for instance 
pathfinding has been traditionally a very active research topic with significant impact on 
the video-game industry, others such as execution monitoring are typically treated in per 
case basis as part of the implementation of the underlying game engine or the particular
character in question. As NPCs become more of real autonomous entities in the game, 
handling such components in a more principled way becomes very important. This is relevant
both from the designers and developers point of view that benefit from an architecture 
that is easier to maintain and reuse, but also from the point of view of the more refined 
interactions with the game-world that NPCs can achieve.

In this paper we propose an architecture for specifying the interaction of NPCs in the 
game-world in a way that abstracts common tasks in four main conceptual components, 
namely \emph{perception, deliberation, control, action}. The architecture is inspired by 
AI research on autonomous agents and robots, in particular the notion of \emph{high-level
control} for \emph{cognitive robotics} as it is used in the context of deliberative
agents and robots such as \cite{levesque98highlevel,Shanahan01Highlevel}. 
The motivation is that by adopting a clear role for each component and specifying a simple 
and clean interface between them, we can have several benefits as we discuss next. 

First, as there are many techniques for specifying how an NPC decides on the next action 
to pursue in the game world, an architecture that abstracts this part in an appropriate 
way could allow to easily switch between approaches. For example, it could facilitate 
experimentation with a finite-state machine \cite{Rabin02FSM}, a behavior tree
\cite{Isla05BehaviorTrees} or a goal-oriented action planning approach \cite{orkin06fear} 
for deciding on NPC actions, in a way that keeps all other parts of the architecture 
agnostic to the actual method for deliberation.

Second, this can provide the ground for a thorough investigation among the different ways
of combining the available methodologies for different components, possibly leading 
to novel ways of using existing approaches. Also, this clear-cut separation of roles can 
encourage the development of modules that encapsulate existing approaches that have not 
been abstracted out of their application setting before, increasing re-usability of 
components.

We also believe that this type of organization is a necessary prerequisite for enabling 
more advanced behaviors that rely on each NPC holding a \emph{personalized view} of the 
game-world that is separated from the current (updated and completely specified) state 
of affairs. In particular, we argue that it is important for believable NPCs to adopt a 
high-level view of relevant aspects of the game-world including the topology and 
connectivity of available areas in the world. In such cases then, when the deliberation 
is more tightly connected with low-level perception and action, we believe that a 
well-principled AI architecture becomes important for maintaining and debugging the 
development process.

For example, typically the game engine includes a pathfinding module that is used by all
NPCs in order to find their way in the game world. But what happens when one NPC knows 
that one path to the target destination is blocked while another NPC does not possess this
information? The approach of handling all requests with a single pathfinding module cannot 
handle this differentiation unless the pathfinder takes into account the personalized 
knowledge of each NPC. This type of mixing perception, deliberation, and action can
be handled in the low-level of pathfinding using various straightforward tricks, but 
making a separation of the high-level knowledge for each NPC and the low-level game-world
state can be very useful.

Adopting this separation, each NPC may keep a personalized high-level view of the 
game-world in terms of available areas or zones in the game and their connectivity, and
use the low-level pathfinding module as a service only for the purpose of finding paths 
between areas that are \emph{known to the NPC} to be connected or points in the same area. 
A simple coarse-grained break-down of the game-world in large areas can ensure that the 
deliberation needed in the high-level is very simple and does not require effort that is
at all similar to the low-level pathfinding. Alternatively, a more sophisticated approach
would be to model this representation based on the actual one used by a hierarchical 
pathfinding approach such as \cite{Botea04HPAstar}, so that the high-level personalized 
deliberation is also facilitated by the pathfinding module but using the NPCs version of
the higher level of the map.

The rest of the paper is organized as follows. We continue  with the 
description of our proposed \emph{CogBot} architecture. We then introduce a motivating 
example that shows the need for personalized knowledge of the game-world for NPCs, and
report on an implementation of CogBots in the popular game engine of 
Unity.\footnote{\url{www.unity3d.com}} 
Then we continue with a discussion on the 
state-of-the-art for approaches that deal with AI for NPCs wrt to actions in the 
game-world, and finally, we conclude with our view on what are 
interesting future directions to investigate. 

\section{The CogBot NPC AI architecture}

The proposed architecture formalizes the behavior of NPCs as far as their actions in the 
game-world is concerned, in terms of four basic components as follows. 

\medskip
\centerline{\includegraphics[width=0.95\linewidth]{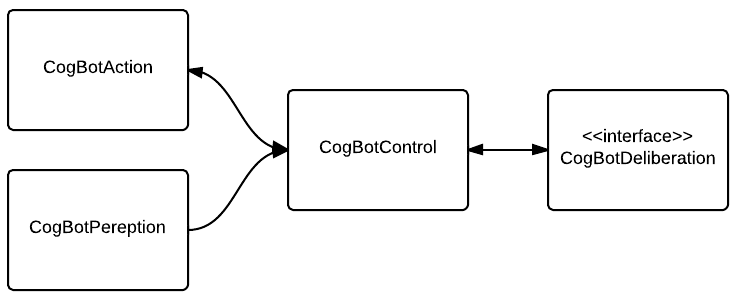}}
\smallskip

\begin{itemize}
\item The \emph{perception} component (\emph{CogBotPerception}) is responsible for 
identifying objects and features 
of the game-world in the field of view of the NPC, including conditions or events that 
occur, which can be useful for the deliberation component. 
\item The \emph{deliberation} component (\emph{CogBotDeliberation}) is responsible for 
deciding the immediate action that should be performed by the NPC by taking into account
the input from the perception component as well as internal representations. This component 
may be used to abstract the logic or strategy that the NPC should follow which could be 
for instance expressed in terms of reactive or proactive behavior following any of 
the existing approaches.
\item The \emph{control} component (\emph{CogBotControl}) is responsible for going 
over a loop that passes information between the perception and deliberation components,
and handling the execution of actions as they are decided. In particular, the controller is 
agnostic of the way that perception, deliberation and action is implemented, but is responsible
for coordinating the information between the components while handling exceptions, monitoring 
conditions and actions, and allocating resources to the deliberator accordingly.
\item The \emph{action} component (\emph{CogBotAction}) is responsible for realizing the 
conceptual actions that are decided by the deliberator in the game-world and provide 
information about the state of action execution, e.g., success or failure.
\end{itemize}

Essentially, the control component acts as a mediator that distributes information between
the other components. Note that more than one instance of each component may be used in an
NPC architecture. This particular specification of components should be seen as a means for
structuring the number of various processes that need to operate and coordinate so that an 
NPC can perform challenging tasks in a game-world environment. One may think of cases where
a single controller is used as a hub that manages many perception, deliberation, and action
components, or other cases where networks of one instance of each these four components are
used to mange different aspects of the NPC behavior.

\subsection{Perception}

The perception component is the main information source for the NPC control component. In 
the typical case it is attached to a mesh object surrounding the NPC and it provides instant
information about all the game objects that lie in the area of the mesh object, e.g., a 
sight cone positioned on the head of the NPC. Also, the perception component may be 
monitoring the field of view for conditions or events which are also propagated to the
control component.

The communication with the control component is asynchronous as the perception component
pushes information to the control component by calling appropriate callback functions as
follows. 

\begin{itemize}
\item An ``Object Entering FoV'' and ``Object Leaving FoV'' callback function is 
called when an object enters or leaves the field of view of the NPC.
\item A ``Notify Object Status'' is called when the internal state of an object in the 
field of view is changed. 
\item A ``Notify Event'' callback function is called whenever an implemented monitoring
condition check is triggered.
\end{itemize}

\medskip
\centerline{\includegraphics[width=.95\linewidth]{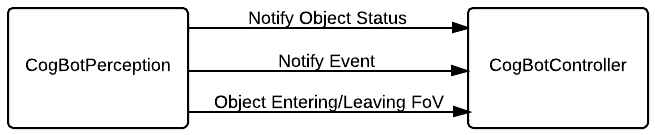}}
\medskip

Moreover, a the perception component provides a filtering mechanism based on the type of
objects and conditions so that only the ones for which the NPC registers for will be reported 
and others will be ignored. In general an NPC may have more than one instance of a perception 
component, each of which may have a different range and can be used to track different 
object types. A simple example is an NPC having one perception mesh for sight and one 
for hearing, which are both set up to communicate with the same control component.

Finally, note that conditions that may be triggered in the environment are propagated as
notification of events. This type of information is also used in other parts of the 
communication between components. This is adopted as a means to provide a simple uniform
mechanism for communication between components.

\subsection{Deliberation}

The deliberation component is the bridge between the low-level space of perception and
action in the game-world and the high-level space of conceptual representation of the
state of affairs as far as the NPC is concerned. The deliberation component exposes an
interface for the control component to asynchronously invoke communication as follows. 
\begin{itemize}
\item A ``Get Next Action'' function abstracts the decision by the deliberation component
with respect to the next immediate action to be performed by the NPC.
\item A ``Notify Object'' function notifies the deliberation component about relevant 
objects that become visible or have changed their state.
\item A ``Notify Event'' function notifies the deliberation component about relevant 
conditions received by the perception component that may affect the internal 
representation of the state or knowledge of the NPC.
\end{itemize}

\medskip
\centerline{\includegraphics[width=0.95\linewidth]{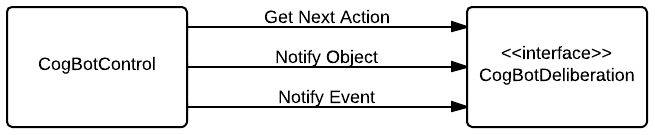}}
\medskip

The deliberation component is responsible for modeling the decision making of the 
NPC with respect to action executing in the game-world, using the information that
is provided by the control component and other internal representations that are
appropriate for each approach or specific implementation.

In particular, one useful view is to think of the deliberation component as maintaining
an internal model of the game-world that is initiated and updated by the low-level input 
coming from the perception component, and using this model along with other models of 
action-based behavior in order to specify the course of action of the NPC. The actual
implementation or method for the model of the game-world and the model of desired 
behavior is not constrained in any way other than the type of input that is provided 
and the type of action descriptions that can be passed to the action component.

Under this abstraction an NPC may for example just keep track of 
simple conditions in the game-world and a representation of its internal state in
terms of health and inventory, and use a finite-state machine approach to specify
the immediate next action to be performed at any time. Similarly, but following a
totally different methodology, an NPC may keep track of a propositional logic
literal-based model of the current state of the game-world, and use a automated planning
decision making process for deciding on the next immediate action to be performed as
part of a longer plan that achieves a desired goal for the NPC.

Observe that in the general case this level of abstraction allows the decision making of
the NPC to possess information that is different that the true state of affairs in the 
game-world. This may be either because some change happened for which the NPC was not
informed by the perception component, for example because the NPC was simply not able
to perceive this information, or even because the perception component itself is 
implemented in such way as to provide filtered or altered information, serving some 
aspect of the game design and game-play.

\subsection{Control}

As we mentioned earlier, the control component acts as a mediator that distributes 
information between the other components, including notifications about the state of 
objects, notifications about conditions in the game-world, as well as feedback about 
action execution. In order to do so it propagates a callback invocation from one 
component to the other (as for example in the case of the perception and deliberation 
component). The way in which this is performed depends on the particular
case and implementation choices. For example different types of information may be 
delivered to different instances of the same component as we discussed earlier in 
this section.

Also, the controller component goes over a loop that handles action execution. In its
most simple form this could be just repeatedly calling in sequence the corresponding
function of the deliberation component that informs about the next action to be
performed, and then pass on this information to the action component so that the
action is actually executed in the game-world. The architecture does not limit or 
prescribe the way that this loop should be implemented and indeed should be done in
a variable way depending on the characteristics of the game. Nonetheless, the intention
is that under this abstraction the control component can encourage more principled 
approaches for action execution monitoring, handling exceptions, and reviving from 
errors.

\subsection{Action}

The action component abstracts the actions of the NPC in the game-world, allowing the 
rest of the architecture to work at a symbolic level and the other components be
be agnostic as per the implementation details for each action. Note that the architecture
view we adopt does not prescribe the level of detail that these actions should be 
abstracted to. For some cases an action could be an atomic low-level task in the game-world
as for example the task of turning the face of the NPC toward some target, while at other
cases a more high-level view would be appropriate, essentially structuring NPC action 
behavior in terms of strategies or macro-actions. In the latter case, the action component
may be used for example to connect a conceptual high-level view of low-level 
implementations of behaviors as it would be typically done when a finite-state machine
is used for reactive behavior.

In terms of the communication of the action component with the control component, again
a very simple interface is adopted for asynchronous interaction as follows.
\begin{itemize}
\item An ``Invoke Action'' function abstracts the action execution from the point of view
of the architecture, initiating an implemented internal function that operates in the 
game-world.
\item A ``Notify Event'' callback function is called to inform the control component about
information related to action execution, such as that the action has finished with success
or that some error occurred, etc.
\end{itemize}
 
\medskip
\centerline{\includegraphics[width=0.95\linewidth]{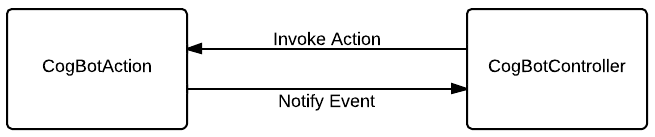}}
\medskip

As in the deliberation component an internal representation is assumed to model the
personalized view of the game-world for the NPC, in the action component a similar 
conceptual representation needs to be utilized in order to capture the available actions
and their characteristics. As in this case the representation can be more straightforward,
we also assume the following simple schema for registering actions in the architecture.
An new action can be registered by means of calling an internal ``Register Action'' 
function which requires i) a string representing the action name (for instance move-to, 
pick-up, etc, and ii) a reference to a function that implements the action. An appropriate
account for additional parameters for actions is needed but this is an implementation 
detail.

We now proceed do discuss a concrete scenario that shows how the CogBot architecture and an
appropriate separation between the low-level state of affairs in the game-world and a personalized
conceptual view of the game-world can enable novel behaviors for NPCs.

\section{A motivating example}

As a motivating running example we will consider a simple case of a room-based 
game-world in which the human player and some NPCs can move between rooms, pickup and 
drop objects, and use some of them to block pathways or set traps. Suppose also that the 
goal of the game is to eventually get hold of a special item and deliver it at the a 
designated spot. 

Now consider the scenario according to which the player decides to block a passage that 
works as a shortcut to some room, by putting some obstacle in the way after he passes 
through. This means that the characters will have to go round using a different route in
order to get close to the human player. This is an important piece of information that
actually should affect greatly how the NPCs will decide to move in the game-world. 

For example, at first, no NPC knows that the passage is blocked so perhaps an NPC that 
wants to go to the room in question will have to go through the normal (shortest) route 
that goes through the shortcut, see that it is blocked, and then go round using the 
alternative route. From then on and until the NPC sees or assumes that the passage is 
cleared, this is what the NPC would be expected to do in order to act in a believable way, 
and this line of reasoning should be adopted for each one of the NPCs separately.

How would this behavior be realized in a game though? This simple scenario suggests that
the pathfinding module of the game-engine should somehow keep track of which obstacles 
each NPC is aware and return a different \emph{personalized path} for each one that is 
consistent with their knowledge. To see why simpler ways to handle this would hurt the 
believability of NPCs consider that the pathfinding module follows an NPC-agnostic 
approach for finding a route such that i) does not take into account object obstacles 
and ii) does take into account object obstacles. In both cases, NPCs are assumed to 
replan if the planned route turns out to be unrealizable.

It is easy to see that both approaches are problematic. In the first case, an NPC may
try to go through the shortcut more than once, in a way that exposes that they has no
way of remembering that the shortcut is blocked. In the second case, an NPC that did 
not first try to use the shortcut will nonetheless immediately choose to follow the 
alternative route even though they did not observe that the shortcut is blocked, probably
ruining the player's plan to slow down other NPCs. 

Essentially, in order to maintain 
the believability of NPCs, each one should be performing pathfinding based on the 
information they possess which needs to be updated accordingly from their observations.
Note how this account of knowledge of topology sets the ground for other more advanced
features for NPCs as for example the capability of exchanging information in the sense
that if a player sees that NPCs gathered together and talked his little trick is no
longer able to slow down other NPCs any more.

We now continue to discuss an application of the CogBot architecture in an implemented
version of this scenario, which is able to handle the intended behavior for NPCs 
making use of an appropriate separation of the actual state of the game-world and
the personalized view of the NPCs. 

\section{CogBots in GridWorld}

\emph{GridWorld} is a grid-based environment developed in the Unity game engine modeled after the 
motivating example we introduced in the previous section. It was specifically designed to be used
for performing automated tests with respect to NPC action-based behavior, therefore all of the
components of the game-world are procedurally generated from an input text file that specifies
the topology as well as the available objects and their state. For example we can design a map 
consisting of rooms interconnected through doors and use it to observe the behavior of one (or 
more) NPCs navigating in the environment while the user opens or closes the doors. In particular
we will experiment with a prototype implementation of our proposed architecture.

\subsection{GridWorld}

The main component of GridWorld is the \emph{map} generated from an ASCII file that is taken as
input. Each element in the grid including wall sections, floor tiles, and doors, is a low-level
game-object of Unity with a component containing some basic information about the object such as:
a flag that indicates whether the object is static or not and its type. Non-static objects, that 
is, objects that may be in more than one states, have an additional component for handle the
internal state representation and the way it changes by means of interactions in the game-world.

A simple A* heuristic search path-finding procedure is implemented based on the grid representation 
in order to provide a low-level navigation system. Moreover, in the initialization phase a processing 
of the map takes place that decomposes the map into interconnected areas using the standard method of 
connected-component labeling \cite{Dillencourt92Component} from the area of image processing. The 
resulting high-level topology is represented as:
\begin{itemize} 
\item A list of \emph{areas}. An area is a fully connected set of tiles in which the movement from 
any tile to any tile is guaranteed to succeed.
\item A list of \emph{way-points} between areas. In our case these way-points are explicitly
represented by \emph{doors}. Each door connects two adjacent areas and has an internal state that
can be open or close.
\item A list of \emph{points of interest} that can be used to model the tiles in the map that are
potential target destinations for the NPC with respect to the scenario in question.
\end{itemize}

This type of information will be maintained by our prototype CogBot NPC, which as we will see 
shortly, combined with the low-level pathfinding procedure can address the challenges raised by
the motivating example we introduced earlier.
The map data ground truth is stored in a global object in the root of the game scene graph. This 
object can be accessed by any other object in the game to ask for world information such as the
size of the map, the object type in some $(i,j)$ position, etc.

\subsection{CogBots}

A prototype NPC in GridWorld is implemented following the CogBot architecture described in
the previous section. The CogBot NPC consists of a standard game character object with a
mesh, colliders, and animations. In addition to these the NPC has the following components:

\begin{itemize}
\item A \emph{conic collider} representing the field of view of the NPC. This collider is attached 
to the main body of the NPC positioned in such way as to simulate its sight cone.
\item A \emph{CogBotPerception} component attached to the conic collider. The perception component
is implemented so as to pass information about all visible objects in the field of view of the NPC.
In this example, no conditions are raised by means of events by the perception component.
\item A \emph{CogBotController} component that simply acts as a bridge for communication between
perception, deliberation, and action.
\item A \emph{PlayerAction} component. This component is a collection of actions of the human
player/spectator in GridWorld that instruct the NPC to perform some activity. 
In this simple example we only have actions from the PlayerAction component that invoke
moving actions for the NPC in order to reach a target destination tile.
\item A \emph{CogBotAction} component that implements the moving actions of the NPC.
\item A \emph{ManualDeliberator} component.  This is an implementation
of the CogBotDeliberator interface that provides the immediate action to be performed by the NPC
based on an internal representation and an model for activity that we will describe with an example
next.
\end{itemize}

\subsection{A simple example}

Consider the following setting for the map of GridWorld.\footnote{
Our prototype implementation of the GridWorld test-bed and the CogBots architecture are 
available at \url{https://github.com/THeK3nger/gridworld} and
\url{https://github.com/THeK3nger/unity-cogbot}. The simple example reported here can 
be found in the ``KBExample'' folder in the GridWorld repository.}

\medskip
\centerline{\includegraphics[width=0.8\linewidth]{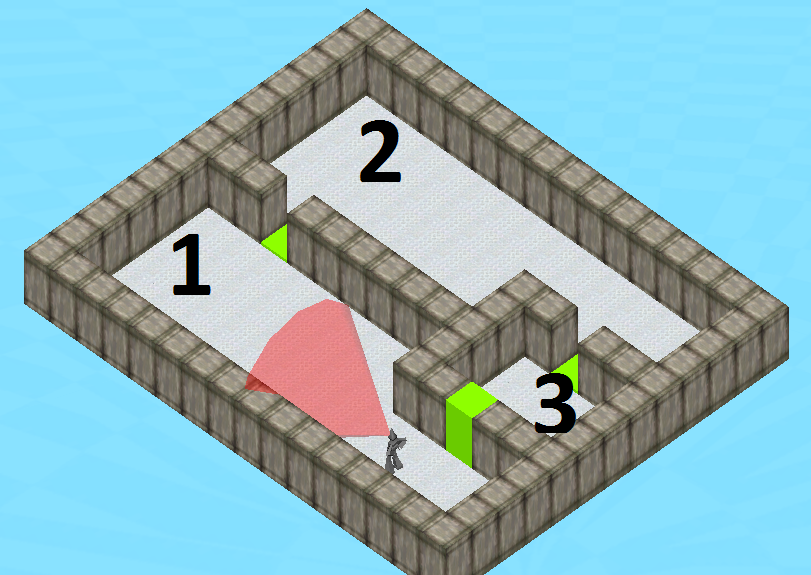}}
\medskip

After the decomposition of the map we have three areas: area 1 where the NPC is located, area 2 that is
symmetrical to area 1 connected through a door, and area 3,  a little room that intervenes between areas 
1 and 2 and connects to both with a door. In particular we call $D_{1,2}$ the door connect areas 1 and 2, and
$D_{1,3}$, $D_{3,2}$ the doors that connect the little room depicted as area 3 with the other two areas.

In this example the human player/observer can instruct the NPC to move to particular tiles by clicking on
them. Suppose for example that the player asks the NPC to go from the initial position in area 1 into area 
2 in a symmetrical position. The actual path to follow would be different depending on which doors are open
allowing the NPC to go through. 

Similarly, assuming an internal representation for the connectivity of 
 areas, and a personalized internal representation of the state of the doors, the NPC could
\emph{deliberate in the level of areas} about the path to take. In other words, the NPC can first build a
coarse-grained high-level plan of the form $\{$``move to area X'', ``move to area Y''$\}$, and then can use
the low-level pathfinding procedure to generate paths for each part of the plan. Apart from achieving a
hierarchical form of pathfinding that could be beneficial for various reasons, this approach actually 
handles the type of believability property that we discussed in the motivating example.

At the beginning of the simulation the internal knowledge of the NPC assumes all doors to be closed.
In this situation instructing the NPC to move to area 2, the deliberation component returns no plan as 
according to the NPC's knowledge it is impossible to go there as all doors are closed.
If we open the doors between 1 and 2 by clicking on them, the deliberation component is still unable to
find a plan because the NPC was not able to get this information through the perception component and its
field of view. Even though the pathfinding procedure would return a valid path, this is irrelevant for the
personalized view of the NPC.

Now assume that we move the NPC close to the doors and the internal representation is updated to reflect
the real world state and take it back to the starting point. If we instruct the NPC to move to area 2, the 
deliberation component produces the straightforward plan which is executed using the pathfinding procedure. 
Similarly, if we open the door $D_{1,3}$, the deliberate component would still return the same plan, as in
order to get the shortest path that goes through area 3 the NPC needs to see that the door is open.

\section{Challenges and related work}

There is a variety of work that aims for action-based AI for NPCs.
Traditionally, a combination of scripts and finite state machine are used in interactive games for controlling NPCs. These methods, even if fairly limited, allow the game designer to control every aspect of the NPCs actions. This approach has been employed in different types of succesfull videogames such as the Role Playing Game (RPG) \textit{Never Winter Nights} or the First Person Shooter (FPS) \textit{Unreal Tournament}. Scripts are written off-line in a high-level language and are used to define simple behaviors for NPCs. Procedural script generation has been proposed in \cite{Mcnaughton04scriptease:generative} by using simple patter templates which are tuned and combined by hand. Complex behaviors can be developed in a short amount of time but many of the intricacies of the classical scripting approach are not solved and it remains difficult to manage the NPCs as the complexity of the virtual world increases.

State machines are still the preferred way to control NPCs in modern games, from FPS like the \textit{Quake} series to RTS like Blizzard's \textit{Warcraft III}. A problem with state machines is that they allow little reusability and they must often be rebuilt for every different case \cite{Orkin2003}. Furthermore, the number of the states grows exponentially if the behavior of the character becomes slightly more sophisticated which is problematic in the design, maintenance, and debugging of NPC behavior.
In Bungie's \textit{Halo 2}, a form of hierarchical state machines or behavior trees are used \cite{Isla05BehaviorTrees}.
In the FPSs from Monolith \textit{F.E.A.R.} and Epic's \textit{Unreal Tournament}, STRIPS and HTN planning have been used to define complex behaviors such as NPCs able to coordinate as squads and perform advanced strategies as sending for backup or flanking \cite{orkin06fear}.
In Bethesda's \textit{Elder Scrolls IV Oblivion}, NPCs are controlled by  
goals that involve scheduling and are given for NPCs to achieve. 
This allows to define behaviors and tasks that depends on preconditions and scheduled times. 
Nonetheless, there is much less effort on standardizing approaches for action-based
AI in a framework that would allow better comparison or collaboration between the existing
techniques. To that end, our proposed architecture allows to:
\begin{itemize}
\item Try out different existing approaches and decision algorithms by abstracting them in the 
as different deliberation compoments that can be easily switched, allowing a comparative AI 
performance analysis.
\item Build NPC that are able to dynamically change the underlying AI method depending on
the state of the game or the vicinity of the player.
For example a simple FSM can be used while an NPC is idle or far away from the player,
and then switch to BT or GOAP when it must defend a location or attack the player.
\item Build NPC that are able to use a combination of existing decision algorithms. For example
we could think of a system that combines BTs for low level decisions and GOAP for high level 
tactical reasoning. The high level planner then would return a sequence of actions, each of 
which would invoke a corresponding BT.
\item Develop NPCs with a \emph{personalized} conceptual representation of the game-world that
enables rich and novel behaviors. The motivating example we examined is just a very simple 
case which to the best of our knowledge has not been handled by approaches in the literature.
Moreover this view can lead to more interesting cases involving also communication between NPC 
and their personalized knowledge.
\end{itemize}

\section{Conclusions}

In this paper we have introduced a robotics-inspired architecture for
handling non-player character artificial intelligence for the purposes of specifying
action in the game world. We demonstrated that certain
benefits can come out of the principled approach for decomposing the artificial 
intelligence
effort in components, in particular with respect to a separation between the 
low-level ground truth for the 
state of the game-world and a personalized conceptual representation of the world for each NPC.

Our proposed architecture provides modularity allowing each of the four main 
components of the architecture, namely, perception, deliberation, control, and
action, to encapsulate a self-contained independent functionality through a clear
interface. We expect that this way of developing characters can enable better code 
reusability and speed up prototyping, testing and debugging.
Moreover our proposed architecture provides the ground for revisiting problems and
techniques that have been extensively studied, such as pahtfinding, and arrive to
feasible methods for developing believable characters with their own view of the
topology and connectivity of the game-world areas.

\clearpage

\bibliographystyle{aaai}
\bibliography{cogbots}

\end{document}